\documentclass{article}
\PassOptionsToPackage{numbers, compress}{natbib}
\usepackage[preprint]{neurips_2022}

\usepackage[utf8]{inputenc} 
\usepackage[T1]{fontenc}    
\usepackage[pagebackref=true,breaklinks=true,colorlinks,citecolor=citecolor, linkcolor=linkcolor]{hyperref}
\usepackage{url}            
\usepackage{booktabs}       
\usepackage{amsfonts}       
\usepackage{amsmath}
\usepackage{nicefrac}       
\usepackage{microtype}      
\usepackage[table]{xcolor}  
\usepackage{tabularx}
\usepackage{graphicx}
\usepackage{multirow}
\usepackage{enumitem}
\usepackage{pifont}
\newcommand{\cmark}{\ding{51}}%
\definecolor{citecolor}{HTML}{0071BC}
\definecolor{linkcolor}{HTML}{ED1C24}

\title{Towards Unifying Medical Vision-and-Language Pre-training via Soft Prompts}

\author{%
  \vspace{0.4em}
  \centerline{
  Zhihong Chen$^{1,2}$$^{*}$ \quad
  Shizhe Diao$^{3}$$^{*}$ } \\
  \textbf{
  Benyou Wang$^{1,2}$$^{\dag}$ \quad
  Guanbin Li$^{4}$$^{\dag}$ \quad
  Xiang Wan$^{2}$} \vspace{0.6em} \\
  \centerline{$^{1}$The Chinese University of Hong Kong, Shenzhen ~ $^{2}$Shenzhen Research Institute of Big Data}\\
  \centerline{$^{3}$The Hong Kong University of Science and Technology ~ $^{4}$Sun Yat-sen University} \\
  \centerline{zhihongchen@link.cuhk.edu.cn ~ sdiaoaa@connect.ust.hk}\\
  \centerline{wangbenyou@cuhk.edu.cn ~ liguanbin@mail.sysu.edu.cn ~ wanxiang@sribd.com}\\
}

\begin{document}

\maketitle

\def\thefootnote{*}\footnotetext{Equal contributions.}
\def\thefootnote{\dag}\footnotetext{Corresponding authors.}
\def\thefootnote{\arabic{footnote}}

\begin{abstract}
Medical vision-and-language pre-training (Med-VLP) has shown promising improvements on many downstream medical tasks owing to its applicability to extracting generic representations from medical images and texts.
Practically, there exist two typical types, \textit{i.e.}, the fusion-encoder type and the dual-encoder type, depending on whether a heavy fusion module is used.
The former is superior at multi-modal tasks owing to the sufficient interaction between modalities; the latter is good at uni-modal and cross-modal tasks due to the single-modality encoding ability.
To take advantage of these two types, we propose an effective yet straightforward scheme named PTUnifier to unify the two types.
We first unify the input format by introducing visual and textual prompts, which serve as a feature bank that stores the most representative images/texts.
By doing so, a single model could serve as a \textit{foundation model} that processes various tasks adopting different input formats (\textit{i.e.}, image-only, text-only, and image-text-pair).
Furthermore, we construct a prompt pool (instead of static ones) to improve diversity and scalability.
Experimental results show that our approach achieves state-of-the-art results on a broad range of tasks, spanning uni-modal tasks (\textit{i.e.}, image/text classification and text summarization), cross-modal tasks (\textit{i.e.}, image-to-text generation and image-text/text-image retrieval), and multi-modal tasks (\textit{i.e.}, visual question answering), demonstrating the effectiveness of our approach.
Note that the adoption of prompts is orthogonal to most existing Med-VLP approaches and could be a beneficial and complementary extension to these approaches.\footnote{Work in progress. The code will be released at \url{https://github.com/zhjohnchan/PTUnifier}.}
\end{abstract}
\section{Introduction}
Medical data is multi-modal in general, among which vision and language are two critical modalities.
It includes visual data (\textit{e.g.}, radiography, magnetic resonance imaging, and computed tomography) and textual data (\textit{e.g.}, radiology reports, and medical texts).
More importantly, such images and texts are pair-collected in routine clinical practice (\textit{e.g.}, X-ray images and their corresponding radiology reports).
Medical vision-and-language pre-training (Med-VLP) aims to learn generic representation from large-scale medical image-text pairs and then transfer it to various medical tasks, which is believed to be beneficial in addressing the data scarcity problem in the medical field.

\begin{figure}[t]
  \centering
  \includegraphics[width=0.95\textwidth]{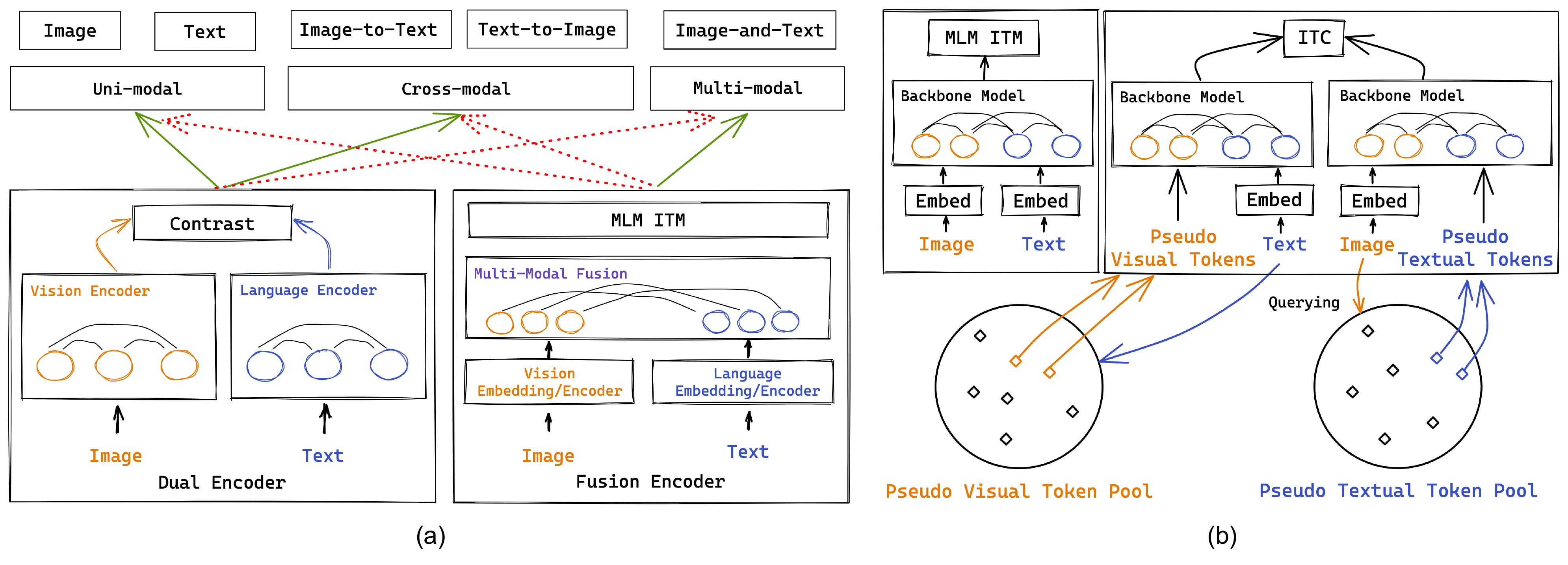}
  \caption{(a) Illustrations of two  Med-VLP paradigms and their advantages (pointed by green arrows) and disadvantages (pointed by red arrows) in downstream tasks; (b) The overall architecture of our proposed approach, where the backbone models share the same parameters, and we duplicate them for illustration.}
  \label{fig:illustration-and-framework}
\end{figure}

Recently, substantial progress has been made toward research on Med-VLP~\cite{zhang2020convirt,li2020comparison,huang2021gloria,moon2021multi,m3ae}.
In general, most existing Med-VLP models can be classified into two types: the dual-encoder type and the fusion-encoder type, where the former encodes images and texts separately to learn uni-modal/cross-modal representations following a shallow interaction layer (\textit{i.e.}, an image-text contrastive layer), and the latter performs an early fusion of the two modalities through the self-attention/co-attention mechanisms to learn multi-modal representations.\footnote{Although the terminologies ``cross-modal'' and ``multi-modal'' have been used interchangeably in the literature, we treat them as terms with different meanings in this paper.}
For dual-encoders, the purpose of existing studies~\cite{zhang2020convirt,huang2021gloria,muller2021lovt,wang2022medclip,wangmulti,wu2023medklip,bannur2023learning} is to develop label-efficient algorithms to learn effective uni-modal/cross-modal representations since large-scale manually labeled datasets are difficult and expensive to obtain for medical images.
The learned representations can improve the \textit{effectiveness} of uni-modal (\textit{i.e.}, vision-only or language-only) tasks\footnote{It is worth noting that most existing studies only conduct the evaluation on the vision-only tasks and disregard the language-only tasks although the text representations are simultaneously learned.} and the \textit{efficiency} of cross-modal (\textit{i.e.}, image-to-text or text-to-image) retrieval tasks significantly.
For fusion-encoders, existing studies~\cite{li2020comparison,khare2021mmbert,moon2021multi,m3ae,arl} aim to jointly process these two modalities with an early interaction to learn multi-modal representations to solve those tasks requiring multi-modal reasoning (\textit{e.g.}, medical visual question answering and medical image-text classification).
However, it seems that ``\textit{you can't have your cake and eat it, too.}'': the fusion-encoders can not perform uni-modal tasks effectively and cross-modal tasks efficiently due to the lack of single-modal encoding, while the dual-encoders underperform on multi-modal tasks owing to the insufficient interaction between modalities as shown in Figure~\ref{fig:illustration-and-framework}(a).

In this paper, we aim to learn a unified medical vision-and-language pre-trained model.
Although there exist some solutions~\cite{wang2021vlmo,singh2022flava} to achieve a similar goal in the general domain, we propose an architecture- and task-agnostic approach named PTUnifier, which is much simpler and lighter-weighted.
Technically, we develop the designs from the following perspectives:
(i) \textit{Compatibility}: we introduce visual and textual prmopts to make the Med-VLP model compatible with different kinds of inputs (\textit{i.e.}, image-only inputs, text-only inputs, and image-text pairs);
(ii) \textit{Scalability}: we improve the diversity of the prompts by constructing prompt pools for different modalities from which different inputs are able to select their corresponding prompts, which enhances the capacity and makes it scalable to larger-scale Med-VLP.
As a result, the proposed approach can be employed in unifying Med-VLP with many existing VLP model architectures (\textit{e.g.}, classic one~\cite{li2019visualbert} or even a single vanilla Transformer model) and does not require extra modality-dependent architectures, resulting in better applicability.
We perform the pre-training on three large-scale medical image-text datasets, \textit{i.e.}, ROCO~\cite{pelka2018roco}, MedICaT~\cite{subramanian2020medicat}, and MIMIC-CXR~\cite{johnson2019mimic}.
To verify the effectiveness of our approach and facilitate further research, we construct a medical vision-language benchmark including {uni-modal} tasks (\textit{i.e.}, image classification (IC) for vision and text classification (TC) and text summarization (TS) for language), {cross-modal} tasks (\textit{i.e.}, image-to-text retrieval (ITR), text-to-image retrieval (TIR), and image-to-text generation\footnote{Medical image-to-text generation refers to medical/radiology report generation in previous studies~\cite{chen2020r2gen,chen2021r2gencmn}.} (ITG)), and {multi-modal} tasks (\textit{i.e.}, visual question answering (VQA)).
The proposed PTUnifier achieves excellent performance on all datasets, demonstrating its effectiveness.

\section{Related Work}
\paragraph{Vision-and-Language Pre-training (VLP)}
Motivated by the success of the self-supervised pre-training recipe in natural language processing (NLP) (\textit{e.g.}, BERT \cite{devlin2019bert}) and computer vision (CV) (\textit{e.g.}, SimCLR \cite{chen2020simclr} and MoCo \cite{he2020moco}), there has been an increasing interest in developing VLP methods to address a wide range of vision-and-language-related tasks.
In general, VLP methods can be classified into two categories according to the vision-and-language interaction, \textit{i.e.}, dual-encoders and fusion-encoders.
Existing dual-encoder methods can be summarized according to the following aspects: (i) using medium-scale curated image-text data \cite{radford2021clip}, (ii) using large-scale noisy image-text data \cite{jia2021align}, (iii) designing more fine-grained image-text contrast \cite{yao2021filip}, (iv) adopting extra single modal contrastive learning \cite{mu2021slip}.
For fusion-encoder approaches, existing studies can be further categorized with respect to these three perspectives:
(i) Uni-modal encoders: different methods adopt different image features (\textit{e.g.}, region features \cite{li2019visualbert,lu2019vilbert}, patch embeddings \cite{kim2021vilt}, and grid features \cite{huang2020pixel}) and distinct text features (\textit{e.g.}, statistic embeddings \cite{kim2021vilt}, and dynamic embeddings \cite{dou2021meter});
(ii) Multi-modal fusion modules: existing studies adopted the single-stream fusion scheme \cite{su2019vlbert,li2020oscar} or dual-stream fusion scheme \cite{tan2019lxmert,yu2021ernie-vil};
(iii) Pretext tasks: existing studies explore a variety of pre-training tasks, including masked language modeling \cite{li2019visualbert}, masked image modeling \cite{lu2019vilbert,chen2020uniter}, image-text matching \cite{zhang2021vinvl}.
This paper adopts the model architecture of fusion encoders and the pretext tasks from both dual-encoder and fusion-encoder types.

\paragraph{Medical Vision-and-Language Pre-Training (Med-VLP)}
Being one of the applications and extensions of VLP to the medical domain, Med-VLP aims to understand the content of medical images and texts, which can be traced back to \cite{zhang2020convirt} for dual-encoders and \cite{li2020comparison} for fusion-encoders.
For dual-encoders, the follow-up studies \cite{huang2021gloria,muller2021lovt,wangmulti} explored the global-local image-text contrastive learning to capture more fine-grained information among medical images and texts and have achieved state-of-the-art results in the medical image classification task.
For fusion-encoders, \cite{khare2021mmbert,moon2021multi,m3ae} performed pre-training to improve the multi-modal reasoning ability of the vision-and-language models for the downstream task (\textit{e.g.}, Medical VQA).
Besides, \cite{arl} integrated medical knowledge into the pre-training procedure to improve the performance on downstream medical tasks.

\paragraph{Unified Vision-and-Language Pre-training}
To unify the dual and fusion encoders, existing studies mainly adopted/designed specific model architectures to accommodate different pretext tasks.
The most common scheme is to add an extra multi-modal fusion module to the dual encoders and perform the cross-modal pretext task (\textit{i.e.}, image-text contrast) before the fusion and multi-modal pretext tasks (\textit{e.g.}, MLM and ITM) after the fusion \cite{li2021albef,singh2022flava}.
Another line of research~\cite{cho2021unifying, wang2022unifying} resorts to multi-tasking on various downstream supervised vision-language tasks by formulating them as sequence-to-sequence tasks. 
Besides, \cite{wang2021vlmo, wang2022image} proposed a mixture-of-modality experts (MoME) Transformer to unify vision-and-language models by employing a set of modality experts to replace the feed-forward networks (FFN) in the standard Transformer.
More recently, \cite{diao2022prefix} proposed an encoder-decoder generative model learned from prefix language modeling and prefix image modeling.
However, the aforementioned studies are either architecture-dependent or task-dependent, and they perform the unifying through training different parts of the models when applying different types of VLP objectives.
Therefore, it is expected to unify the existing Med-VLP types in an \textit{architecture- and task-agnostic} fashion to improve the generalization and extensionality ability of Med-VLP methods.

\section{Bridging the Gap}
In this section, we introduce the PTUnifier framework for unifying the fusion-encoder and dual-encoder types.
\S\ref{sec:problem} details the problem to be addressed.
This work proposes to unify inputs using prompts (in \S\ref{sec:unify_input}).
Thus one could jointly train various tasks even with different input formats (in \S\ref{subsec:pre-training-objectives}) in either pre-training or fine-tuning.

\subsection{Problem Definition}
\label{sec:problem}
We adopt the general problem formulation for pre-training following existing studies \cite{li2019visualbert,tan2019lxmert}.  
Formally, given a medical image $I$ and its corresponding description text $T$, the representation learning process can be formulated as
\begin{equation}
\label{eq:objective}
    \theta^{*},\theta_{1}^{*},...,\theta_{S}^{*} = \mathop{\arg\min}_{\theta,\theta_{1},...,\theta_{S}}
    \sum_{s=1}^{S} \mathcal{L}_{s}
    (Y_{s}, \mathcal{H}_{\theta_{s}}(\mathcal{M}_{\theta}(\boldsymbol{X})),
\end{equation}
where $S$ refers to the number of pretext tasks; $\mathcal{L}_{s}$ are the loss functions of pretext tasks; $Y_{s}$ are the corresponding ground-truth labels; $\mathcal{H}_{\theta_{s}}$ are the prediction heads with their parameters $\theta_{s}$; $\mathcal{M}_{\theta}$ is the backbone model which is parameterized by $\theta$; $\boldsymbol{X}$ represents the input to the backbone model, which could be one of the following cases:
\begin{equation}
    \boldsymbol{X} = \begin{cases} (\boldsymbol{X}^{v}) & \textrm{if} \quad \textit{image-only} \\
    (\boldsymbol{X}^{l})   & \textrm{if} \quad \textit{text-only}\\
     (\boldsymbol{X}^{v}, \boldsymbol{X}^{l})  & \textrm{if} \quad \textit{image-text} \\
    \end{cases}
\end{equation}
where we suppose that we have embedded a medical image $I$ as $\boldsymbol{X}^{v} \in \mathbb{R}^{D_{v} \times N_{v}}$ or a medical text $T$ as  $\boldsymbol{X}^{l} \in \mathbb{R}^{D_{l} \times N_{l}}$ when dealing with vision and language modalities.
The challenge of the problem is to make the backbone model $\mathcal{M}_{\theta}$  deal with such variable-size and heterogeneous input.
After overcoming this challenge, we can perform different types of downstream vision-language tasks (\textit{i.e.}, uni-modal, cross-modal, and multi-modal tasks).

\subsection{Unifying Inputs via Prompts}
\label{sec:unify_input}
To unify inputs, we propose to unify the inputs via prompts so as to perform different types of tasks.
In specific, we design two solutions, \textit{i.e.}, a basic solution for \textit{compatibility} and an advanced solution for \textit{scalability}.
In this work, we use the advanced solution in default if not specified.

\subsubsection{Compatibility using Prompts}
\label{sec:static}
To make the backbone model compatible with variable-size and heterogeneous input, this work proposes a simple yet effective approach, namely using Prompt (PT) as a placeholder for missing modality. $\mathcal{M}_{\theta}$ naturally accepts two inputs (visual and textual embeddings $\boldsymbol{X}^{v}, \boldsymbol{X}^{l}$), which is by definition compatible to inputs with image-text pairs.
For image-only/text-only inputs, we propose to introduce visual/textual prompts to enable the backbone model to perceive the missing input in a specific modality:
\begin{equation}
\label{eq:pseduo}
    \boldsymbol{X} = \begin{cases} ( \boldsymbol{X}^{v} , {\color{red} \boldsymbol{PT}^{l}}) & \textrm{if} \quad \textit{image-only} \\
    ({ \color{red} \boldsymbol{PT}^{v}},\boldsymbol{X}^{l}) & \textrm{if} \quad \textit{text-only} \\
     (\boldsymbol{X}^{v}, \boldsymbol{X}^{l})  & \textrm{if} \quad \textit{image-text}\\
    \end{cases}
\end{equation}
where $\boldsymbol{PT}^{v} \in \mathbb{R}^{D_{v} \times k}$ and $\boldsymbol{PT}^{l} \in \mathbb{R}^{D_{l} \times k}$ are the visual and textual prompts, respectively.
\subsubsection{Scalability of Prompts}
\label{sec:dynamic}
The above solution adopts a static fashion to introduce prompts, which might have limited diversity and therefore harm its capacity. Hence, we construct a pool of visual/textual prompts \textit{instead of static prompts}. Importantly, the selection of prompts is  \textit{conditioned on the input embeddings}.

Formally, we define a visual prompt pool $\boldsymbol{V} \in \mathbb{R}^{D_{v} \times N_{v}}$ and a textual prompt pool $\boldsymbol{T} \in \mathbb{R}^{D_{l} \times N_{l}}$. $N_{v}$ and $N_{l}$ are the size of the visual/textual prompt pool, respectively.
Given the image-only input with its visual embedding sequence $\boldsymbol{X}^{v}$ or language-only input with its textual embedding sequence $\boldsymbol{X}^{l}$, we conduct a pooling operation (\textit{e.g.}, average/max pooling) to obtain a \textit{query vector} for existing modality (denoted as $\boldsymbol{q}^{v}$ or $\boldsymbol{q}^{l}$), namely,  $\boldsymbol{q}^{v} = \textrm{pooling}(\boldsymbol{X}^{v}) $ and $\boldsymbol{q}^{l} = \textrm{pooling}(\boldsymbol{X}^{l}) $, respectively.
To get the prompts of the missing modality, the selection of prompts is based on the similarity scores between the query vector and all prompts in the pool from the missing modality:
\begin{equation}
\label{eq:selection}
\begin{aligned}
 \boldsymbol{PT}^{l} =\underset{\boldsymbol{w}\in \boldsymbol{V}}{\text{top-}k} \left[{\boldsymbol{w}^{T}\boldsymbol{q}^{v}}\right],\\
    \boldsymbol{PT}^{v} =\underset{\boldsymbol{w}\in \boldsymbol{T}}{\text{top-}k} \left[\boldsymbol{w}^{T}{\boldsymbol{q}^{l}}\right],
\end{aligned}
\end{equation}
where $\boldsymbol{w}$ is an embedding vector in the prompt pool, and we select $k$ closest prompts as the input embedding sequence of the missing modality. 

\paragraph{Intuitive Explaination}
Without loss of any generality, we take a text-only scenario as an example, but it also holds for the image-only scenario. To select the best visual prompts for the text-only input, the proposed method chooses the most similar ones compared to the given textual query vector. As an intuitive explanation, one could treat the visual prompt pool as a feature bank that stores the most representative images of a given dataset, Eq.~\ref{eq:selection} aims to choose the visual prompts that might convey a similar semantic meaning as the given text by conducting dot products. In other words, \textit{it might, at least to some extent, automatically fill (originally unprovided)  semantically-similar images conditioned on purely the given text}.

\paragraph{Linking to Prompts}
We find that the PTUnifier (especially the static one in \S\ref{sec:static}) is quite similar to the prompt tuning \cite{li2021prefix,liu2021pre}. They both introduce special tokens or vectors as a certain signal for training or inference. One notable difference is that in a special version of PTUnifier using prompts pools (see \S\ref{sec:dynamic}),  the selection of additional tokens/vectors is conditioned on the input, while prompts are generally static and constant to input.

\subsection{Unifying Multiple Pre-training Objectives}
\label{subsec:pre-training-objectives}
Owing to the unified image and/or text input formulation, we can adopt pretext tasks of both fusion-encoders and dual-encoders (see Eq.~\ref{eq:objective}).
Following previous studies \cite{li2019visualbert,tan2019lxmert,zhang2020convirt,radford2021clip}, we develop two commonly used pretext tasks (\textit{i.e.}, masked language modeling (MLM) and image-text matching (ITM)) for fusion-encoders and the image-text contrast (ITC) pretext task for dual-encoders.
To produce the prediction for the aforementioned MLM and ITM tasks, we use two independent prediction heads $\mathcal{H}_{\rm MLM}$ and $\mathcal{H}_{\rm ITM}$ (\textit{i.e.}, two two-layer multilayer perceptrons (MLP)).

\paragraph{Masked Language Modeling (MLM)}
Following BERT~\cite{devlin2019bert}, we randomly mask 15\% of the words (denoted as $Y_{\rm MLM}$) of the input text $T$ and recover them according to the remaining text ($T_{\rm M}$) and the input $I$.
The MLM objective is given by:
\begin{equation}
    \mathcal{L}_{\rm MLM} = - \sum_{(I, T)} {\log{p_{\rm MLM}(Y_{\rm MLM} | I, T_{\rm M})}},
\end{equation}
where $p_{\rm MLM}$ is obtained by applying $\mathcal{H}_{\rm MLM}$ followed by a softmax operation on the corresponding representations of \texttt{[MASK]} in $\boldsymbol{Z}^{l}$.

\paragraph{Image-Text Matching (ITM)}
 aims to distinguish whether an image-text pair is a match.
In detail, a positive image-text pair and a randomly sampled negative pair are fed into $\mathcal{M}_{\theta}$ and the concatenation of $\boldsymbol{z}^{v}_{\rm [CLS]}$ and $\boldsymbol{z}^{l}_{\rm [CLS]}$ is processed by $\mathcal{H}_{\rm MLM}$ followed by a softmax layer to output a binary probability $p_{\rm ITM}$.
Therefore, the ITM objective is given by
\begin{equation}
    \mathcal{L}_{\rm ITM} = - \sum_{(I, T)} {\log{p_{\rm ITM}(Y_{\rm ITM} | I, T)}}.
\end{equation} 

\paragraph{Image-Text Contrast (ITC)}
aims to learn better uni-modal/cross-modal representation from the instance-level contrast.
In this work, given an image-text pair, we use two different forward procedures on the image-only input $I$ and the text-only input $T$, respectively, to obtain the image-only representation (denoted as $\boldsymbol{z}^{v}$) and text-only representation (denoted as $\boldsymbol{z}^{l}$).
Afterward, we adopt the similarity function $s\left(I, T\right)={\boldsymbol{z}^{v}}^{\top}\boldsymbol{z}^{l}$ to compute the image-to-text similarity and text-to-image similarity between $\boldsymbol{z}^{v}$ and $\boldsymbol{z}^{l}$.
Subsequently, the similarities are normalized as follows:
\begin{align}
    p_{n}^{\mathrm{i} 2 \mathrm{t}}&=\frac{\exp \left(s\left(I, T_{n}\right) / \tau\right)}{\sum_{n=1}^{N} \exp \left(s\left(I, T_{n}\right) / \tau\right)},\\
    p_{n}^{\mathrm{t} 2 \mathrm{i}}&=\frac{\exp \left(s\left(I_{n}, T\right) / \tau\right)}{\sum_{n=1}^{N} \exp \left(s\left(I_{n}, T\right) / \tau\right)},
\end{align}
where $N$ is the size of the mini-batch.
The ground-truth labels $Y^{\mathrm{i} 2 \mathrm{t}}$ and $Y^{\mathrm{t} 2 \mathrm{i}}$ are two $N \times N$ one-hot matrices, where negative pairs have a probability of 0 and the positive pair has a probability of 1.
Therefore, the ITC objective is given by
\begin{equation}
\mathcal{L}_{ITC} = -\frac{1}{2} \sum_{(I, T)} \log{p^{\mathrm{i} 2 \mathrm{t}}(Y^{\mathrm{i} 2 \mathrm{t}} | I, T)} -\frac{1}{2} \sum_{(I, T)} \log{p^{\mathrm{t} 2 \mathrm{i}}(Y^{\mathrm{t} 2 \mathrm{i}}| I, T)}.
\end{equation}

\section{The Model Architecture}
The previous section documents the unification at the input and task levels. This section will introduce the overall architecture of our work.
As a pipeline, we first map visual and textual tokens into embeddings space ($\boldsymbol{X}^{v}$ and $\boldsymbol{X}^{l}$ as specified in \S\ref{sec:embedding}). Such token embeddings with or without prompts will be jointly processed by an identical backbone model $\mathcal{M}_{\theta}$ (\S\ref{subsec:model-architecture}).
An overview of the proposed approach is shown in Figure~\ref{fig:illustration-and-framework}(b).

\subsection{Visual and Textual Embeddings}
\label{sec:embedding}
\paragraph{Visual embedding}
For an input image $I$, it is first segmented into patches following~\cite{dosovitskiy2020vit}.
Then the patches are linearly projected into patch embeddings $\boldsymbol{X}^{v} = (\boldsymbol{x}_{1}^{v}, \boldsymbol{x}_{2}^{v}, \ldots, \boldsymbol{x}_{N_{v}}^{v}), \boldsymbol{x}_{i}^{v} \in \mathbb{R}^{D_{v}}$ through a linear transformation and a special learnable token embedding $\boldsymbol{x}_{\rm [CLS]}^{v}$ is prepended for the aggregation of visual information.
Therefore, the image embedding sequence is obtained by summing up the patch embeddings and learnable 1D position embeddings $\boldsymbol{E}^{v}_{pos} \in \mathbb{R}^{D_{v} \times (N_{v}+1)}$:
\begin{equation}
    \boldsymbol{X}^{v} = [\boldsymbol{x}_{\rm [CLS]}^{v};\boldsymbol{x}_{1}^{v};\boldsymbol{x}_{2}^{v};...;\boldsymbol{x}_{N_{v}}^{v}] + \boldsymbol{E}^{v}_{pos},
\end{equation}
where $[\cdot;\cdot]$ represents the column concatenation.\footnote{We overload the notation $\boldsymbol{X}^{v}$ for simplicity (same for $\boldsymbol{X}^{l}$).}

\paragraph{Textual embedding}
Similarly, for an input text $T$, we follow BERT \cite{devlin2019bert} to tokenize the input text to subword tokens by WordPiece \cite{wu2016google}.
Afterwards, the tokens are linearly projected into embeddings $\boldsymbol{X}^{l} = (\boldsymbol{x}_{1}^{l}, \boldsymbol{x}_{2}^{l},...,\boldsymbol{x}_{N_{l}}^{l}), \boldsymbol{x}_{i}^{l} \in \mathbb{R}^{D}$ through a linear transformation with a start-of-sequence token embedding $\boldsymbol{x}_{\rm [CLS]}^{l}$, and a special boundary token embedding $\boldsymbol{x}_{\rm [SEP]}^{l}$ added.
Therefore, the text embedding sequence is obtained by summing up the sub-word token embeddings and text position embeddings $\boldsymbol{E}^{l}_{pos} \in \mathbb{R}^{D \times (N_{l}+2)}$:
\begin{equation}
    \boldsymbol{X}^{l} = [\boldsymbol{x}_{\rm [CLS]}^{l}; \boldsymbol{x}_{1}^{l}; \ldots; \boldsymbol{x}_{N_{l}}^{l}; \boldsymbol{{x}}_{\rm [SEP]}^{l}] + \boldsymbol{E}^{l}_{pos}.
\end{equation}

\subsection{The Backbone Model}
\label{subsec:model-architecture}
Since the input image and/or text are represented as a unified image-text sequence, the backbone model can be any model for sequential modeling.
In this work, we adopt an attention-based Med-VLP model with the multi-modal interaction, which can be an effective model (including uni-modal encoders and a multi-modal fusion module) or an efficient one (\textit{i.e.}, a single Transformer model),
where the attention mechanism is defined as
\begin{equation}
    {\rm ATTN}(\boldsymbol{Q},\boldsymbol{K},\boldsymbol{V})=\operatorname{softmax}\left(\boldsymbol{Q} \boldsymbol{K}^{T}/\sqrt{D_{k}}\right) \boldsymbol{V},
\end{equation}
where $\boldsymbol{Q}$, $\boldsymbol{K}$, and $\boldsymbol{V}$ are the query, key, and value matrix linearly transformed from the input embedding sequence, respectively, and $D_{k}$ is the dimension of $\boldsymbol{K}$.
Formally, for a given input (defined in Eq.~\ref{eq:pseduo}), the whole representation process can be formulated as
\begin{equation}
   \boldsymbol{Z}^{v}, \boldsymbol{Z}^{l} = \mathcal{M}_{\theta}(  \boldsymbol{X} ),
\end{equation}
where $\boldsymbol{Z}^{v}=(\boldsymbol{z}_{\rm [CLS]}^{v},\boldsymbol{z}_{1}^{v},\boldsymbol{z}_{2}^{v},...,\boldsymbol{z}_{N_{v}}^{v})$ and $\boldsymbol{Z}^{l}=(\boldsymbol{z}_{\rm [CLS]}^{l},\boldsymbol{z}_{1}^{l},..., \boldsymbol{z}_{N_{l}}^{l},\boldsymbol{{z}}_{\rm [SEP]}^{l})$ are the image and text representations from the backbone model.

\section{Experimental Settings}
\subsection{Pre-training Datasets}
\label{sec:pre-training-datasets-mainbody}
In our experiments, we perform the pre-training on three datasets, which are described as follows:
\begin{itemize}[leftmargin=*,noitemsep,nolistsep]
    \item \textbf{ROCO} \cite{pelka2018roco}: a dataset of radiology figure-caption pairs from PubMed Central, an open-access biomedical literature database.
    \item \textbf{MedICaT} \cite{subramanian2020medicat}: a dataset of medical figure-caption pairs also extracted from PubMed Central.
    Different from ROCO, 75\% of its figures are compound figures, including several sub-figures.
    \item \textbf{MIMIC-CXR} \cite{johnson2019mimic}: the largest radiology dataset to date consisting of chest X-ray images (in frontal or lateral views) and their reports from the Beth Israel Deaconess Medical Center.
\end{itemize}
For all the datasets, we exclude those samples with a text length of less than 3.
For ROCO and MedICaT, we filter non-radiology samples, and for MIMIC-CXR, we only keep images in the frontal view.
As for the dataset split, we adopt the official splits of ROCO and MIMIC-CXR.
For MedICaT, we randomly sample 1,000 image-text pairs for validation and 1,000 for testing, and the remaining image-text pairs are used for training.
Different from the texts in general-domain VLP, medical texts are long narratives consisting of multiple sentences.
To deal with this case, we randomly sample a sentence from the input text in each iteration.

\subsection{Medical Vision-Language  Benchmark}
\label{sec:downstream-datasets}
To evaluate the performance, we construct a medical vision-language evaluation benchmark including three types of tasks, \textit{i.e.}, uni-modal, cross-modal, and multi-modal evaluations.\footnote{More details of the downstream evaluations are reported in Appendix A.}
All the adopted datasets are related to radiology.

\paragraph{Uni-modal Evaluation}
 requires the model to process a single modality with vision-only or language-only inputs.
For vision-only tasks, we conduct the image classification (IC) experiments on CheXpert \cite{irvin2019chexpert} and RSNA Pneumonia \cite{shih2019rnas}.
For language-only tasks, we perform both the understanding task (\textit{i.e.}, text classification (TC)) and the generation task (\textit{i.e.}, text summarization (TS)) on the RadNLI \cite{romanov2018mednli,miura2021ifcc} and MIMIC-CXR datasets, respectively.

\paragraph{Cross-modal Evaluation}
 requires the model to align the vision and language modalities.
We conduct experiments on three kinds of tasks (\textit{i.e.}, image-to-text retrieval (ITR), text-to-image retrieval (TIR), and image-to-text generation (ITG)).
For ITR and TIR, we adopt the ROCO dataset and measure both zero-shot and fine-tuned performance.
During the evaluation, we sample 2,000 image-text pairs from the ROCO test set and report the results on the 2,000 sampled image-text pairs due to the large time complexity of the ranking process.
For ITG, we conduct experiments on the MIMIC-CXR dataset to evaluate its ability for radiology report generation.

\paragraph{Multi-modal Evaluation}
 requires the model to reason over both the image and text inputs through the multi-modal interaction.
We conduct the experiments on the medical visual question answering (VQA) task, which requires the model to answer natural language questions about a medical image.
We adopt three publicly available Med-VQA datasets (\textit{i.e.}, VQA-RAD \cite{Lau2018VQARAD}, SLAKE \cite{liu2021slake}, and MedVQA-2019 \cite{abacha2019medvqa}), where VQA-RAD consists of 3,515 image-question pairs, SLAKE contains 14,028 image-question pairs and MedVQA-2019 contains 15,292 image-question pairs.

\subsection{Implementation Details}
\paragraph{Pre-training}
We adopt the classical VLP model as the backbone model, including a vision encoder, a language encoder, and a multi-modal fusion module.
For the vision and language encoders, we adopt base-size Transformer encoders with 12 layers initialized from CLIP-ViT-B \cite{radford2021clip} RoBERTa-base \cite{liu2019roberta} and their hidden dimension is set to 768.
For the multi-modal fusion module, we set the number of Transformer layers to 6, the dimension of the hidden states to 768, and the number of heads to 12.
For the visual/textual prompt pools, the dimension and the pool size is set to 768 and 1,024, respectively, by default.
For optimization, the pre-training takes 100,000 steps with AdamW optimizer \cite{loshchilov2017decoupled} with a weight decay of 0.01.
The learning rates for the vision and language encoders and the remaining parameters are set to 1e-5 and 5e-5, respectively.
We use the warm-up strategy during the first 10\% of the total number of steps, and the learning rate is linearly decayed to 0 after warm-up.
For data augmentation, we use center-crop to resize each image to the size of 288$\times$288.

\paragraph{Fine-tuning}
For all downstream tasks, we use the AdamW optimizer with the learning rate set to 5e-6 and 2.5e-4 for the backbone model and task-specific layers, respectively.
The fine-tuning strategies can be divided into three categories according to the type of tasks.
Specifically, for the classification tasks (\textit{i.e.}, IC, TC, and VQA), we feed the concatenation of the image/visual prompt and text/textual prompt representations to a randomly initialized two-layer MLP to predict the labels.
For the retrieval tasks (\textit{i.e.}, ITR and TIR), we adopt the prediction head for the image-text contrast pre-text task and test its zero-shot and fine-tuned performance.
For the generation tasks (\textit{i.e.}, TS and ITG), we feed the concatenation of the sequence of image/visual prompt and text/textual prompt representations to a Transformer decoder with its parameters (except for the parameters of cross-attention layers) initialized from the pre-trained language encoder.
For the evaluation metrics, we follow the previous studies to adopt AUROC for IC, accuracy for TC and VQA, Recall@K (K=1, 5, 10) for ITR and TIR, and natural language generation (NLG) metrics (\textit{i.e.}, BLEU \cite{bleu}, METEOR \cite{meteor}, CIDEr \cite{cider}, and ROUGE \cite{rouge}) for TS and ITG.

All pre-training and fine-tuning experiments are conducted on 80GB NVIDIA A100 GPUs with mixed-precision \cite{micikevicius2017mixed} to accelerate training and save memory.
\section{Results and Analyses}
\subsection{Main Results}
\begin{table*}[t]
\footnotesize
\centering
\setlength{\tabcolsep}{0.8mm}{
\resizebox{0.99\linewidth}{!}{
\begin{tabular}{@{}lcccccccccc@{}}
\toprule
                           & \multicolumn{4}{c}{Uni-Modal}                                                                                                 & \multicolumn{3}{c}{Cross-Modal}                                                            & \multicolumn{3}{c}{}                                                                       \\
                           & \multicolumn{2}{c}{Image}                                   & \multicolumn{2}{c}{Text}                                        & \multicolumn{2}{c}{Image-to-Text}                           & Text-to-Image                & \multicolumn{3}{c}{\multirow{-2}{*}{Multi-Modal}}                                          \\ \cmidrule(l){2-11} 
                           & CheXpert                     & PNAS                         & RadNLI                           & MIMIC                        & MIMIC                        & ROCO                         & ROCO                         & VQA-RAD                      & SLAKE                        & MedVQA-2019                  \\
\multirow{-4}{*}{Methods}  & AUROC                        & AUROC                        & Acc                              & RL                           & BL4                          & R@1                          & R@1                          & Acc                          & Acc                          & Acc                          \\ \midrule
                           & \multicolumn{2}{c}{ConVIRT}                                 & ClinicalBERT                     & TransABS                     & R2Gen                        & \multicolumn{2}{c}{ViLT}                                    & \multicolumn{3}{c}{CPRD}                                                                   \\
                           & \multicolumn{2}{c}{\cite{zhang2020convirt}}                 & \cite{alsentzer2019clinicalbert} & \cite{liu2019transabs}       & \cite{chen2020r2gen}         & \multicolumn{2}{c}{\cite{kim2021vilt}}                      & \multicolumn{3}{c}{\cite{liu2021cprd}}                                                     \\
\multirow{-3}{*}{Study$_1$} & 87.3                         & 81.3                         & 72.6                             & 43.8                         & 8.0                          & 11.9                         & 9.8                          & \cellcolor[HTML]{EFEFEF}72.7 & \cellcolor[HTML]{EFEFEF}82.1 & -                            \\ \midrule
                           & \multicolumn{2}{c}{GLoRIA}                                  & IFCC                             & WGSum                        & M2Trans                     & \multicolumn{2}{c}{METER}                                   & \multicolumn{3}{c}{MMBERT}                                                                 \\
                           & \multicolumn{2}{c}{\cite{huang2021gloria}}                  & \cite{miura2021ifcc}             & \cite{hu2021wgsum}           & \cite{miura2021ifcc}      & \multicolumn{2}{c}{\cite{dou2021meter}}                            & \multicolumn{3}{c}{\cite{khare2021mmbert}}                                                 \\
\multirow{-3}{*}{Study$_2$} & \cellcolor[HTML]{EFEFEF}88.1 & \cellcolor[HTML]{EFEFEF}88.6 & \cellcolor[HTML]{EFEFEF}77.8     & \cellcolor[HTML]{EFEFEF}45.1 & \cellcolor[HTML]{EFEFEF}10.5 & \cellcolor[HTML]{EFEFEF}14.5 & \cellcolor[HTML]{EFEFEF}11.3 & 72.0                         & -                            & \cellcolor[HTML]{EFEFEF}77.9 \\ \midrule
PTUnifier (ours)           & \cellcolor[HTML]{C0C0C0}90.1 & \cellcolor[HTML]{C0C0C0}90.6  & \cellcolor[HTML]{C0C0C0}80.0     & \cellcolor[HTML]{C0C0C0}46.2 & \cellcolor[HTML]{C0C0C0}10.7 & \cellcolor[HTML]{C0C0C0}21.0 & \cellcolor[HTML]{C0C0C0}20.8 & \cellcolor[HTML]{C0C0C0}78.3 & \cellcolor[HTML]{C0C0C0}85.2 & \cellcolor[HTML]{C0C0C0}79.3 \\ \bottomrule
\end{tabular}}}
\caption{Comparisons of our proposed method with previous studies on three types of evaluations (\textit{i.e.}, uni-modal, cross-modal, and multi-modal evaluations). 
Study$_{1}$ and Study$_{2}$ denote two state-of-the-art approaches of each type of tasks, respectively.
BL-4 denotes BLEU score using 4-grams and RG-L denotes ROUGE-L (same below).
Dark and light grey colors highlight the top and second best results on each metric (same below).
Note that the results of text summarization and image-to-text generation are replicated using our pre-processed data (See Appendix A).}
\label{table:main-experiments}
\end{table*}
To demonstrate the effectiveness of the proposed approach, we conduct experiments on the aforementioned medical vision-language benchmark.
The results of the main experiments are reported in Table \ref{table:main-experiments}.
There are several observations.
First, our approach achieves the best performance on all tasks.
It outperforms previous studies on uni-modal image classification (+2.0\% AUROC), text classification (+2.2\% Accuracy), text summarization (+1.1\% Rouge-L), image-to-text generation (+0.2\% BLEU-4), image-to-text retrieval (+7.5\% Recall@1), text-to-image retrieval (+9.5\% Recall@1), and multi-modal VQA (+3.4\% Accuracy), which confirms the validity of the proposed approach.
Second, the proposed approach outperforms those complicated methods designed for specific tasks.
For example, R2Gen introduced recurrent memory networks to the Transformer decoder to augment its decoding ability;
WGSum used extra word graphs to improve the ability to detect keywords in the findings section of radiology reports.
CPRD adopted representation distillation to alleviate the data scarcity problem in the Med-VQA task.
These observations show that different pre-training ways can enable distinct abilities of the model, and it is possible to design an appropriate approach to exploit the knowledge shared across different tasks and perform various tasks using a unified model.
Note that \textit{the existing studies are only designed for a single task}, while our approach generally targets all vision- and/or language-related tasks, namely, without any tailored adaptations to a specific task.

\subsection{Ablation Study}
\label{sec:ablation_study}
To further illustrate the effectiveness of our proposed approach, we perform an ablation study on the pre-training objectives, including the ones from fusion-encoders (\textit{i.e.}, MLM and ITM) and the one from dual-encoders (\textit{i.e.}, ITC).

There are several observations drawn from different aspects.
First, the objectives of fusion encoders (\textit{i.e.}, MLM and ITM) guide the models (\textit{i.e.}, ID 3 and 5) to the more powerful multi-modal representations than other models without them, which could be observed from the performance on the downstream Med-VQA task.
Second, the image-text contrast objective of dual encoders assists the models (\textit{i.e.}, ID 4 and 5) in learning the uni-modal image representations and the cross-modal representations, and the models pre-trained with the ITC objective outperform those pre-trained without the ITC objective.
More importantly, the models pre-trained with the ITC objective (\textit{i.e.}, ID 4 and 5) demonstrate their great transfer ability where the pre-trained models can achieve high performance with very little data (\textit{e.g.}, 1\% and 10\%).
Third, it is interesting to note that the ITC objective does not promote the performance of the uni-modal text classification task.
We can explain this phenomenon by the reason that images and texts are abstracted at different levels, where pixels of images have a lower semantic level than tokens of texts.
Therefore, in the ITC process, the texts can be treated as a kind of ``supervision signals'' for the learning of image encoding, yet, it is harder for the images to play such a role in contrast.
This can be observed from previous studies \cite{radford2021clip,jia2021align,mu2021slip}, where the dual-encoders were only evaluated on the uni-modal vision tasks or cross-modal tasks.
Fourth, performing both types of objectives promotes the model (\textit{i.e.}, ID 5) to achieve the best performance across all the tasks, which confirms the feasibility of the research direction on unifying the fusion-encoders and dual-encoders.
\begin{table*}[t]
\centering
\footnotesize
\resizebox{0.99\linewidth}{!}{
\begin{tabular}{@{}lcccccccccccc@{}}
\toprule
                     & \multicolumn{3}{c}{}                                                 & \multicolumn{4}{c}{Uni-Modal}                                                                                             & \multicolumn{2}{c}{Cross-Modal}                             & \multicolumn{3}{c}{}                                                                       \\
                     & \multicolumn{3}{c}{\multirow{-2}{*}{Objectives}}                     & \multicolumn{3}{c}{Image}                                                                  & Text                         & \multicolumn{2}{c}{Image-to-Text}                           & \multicolumn{3}{c}{\multirow{-2}{*}{Multi-Modal}}                                          \\ \cmidrule(l){2-13} 
                     &                       &                       &                      & \multicolumn{3}{c}{CheXpert}                                                               &                              & \multicolumn{2}{c}{}                                        & \multicolumn{3}{c}{VQA-RAD}                                                                \\
                     &                       &                       &                      & 1\%                          & 10\%                         & 100\%                        & \multirow{-2}{*}{RadNLI}     & \multicolumn{2}{c}{\multirow{-2}{*}{MIMIC}}                 & Open                         & Closed                       & Overall                      \\
\multirow{-5}{*}{ID} & \multirow{-3}{*}{MLM} & \multirow{-3}{*}{ITM} & \multirow{-3}{*}{ITC} & AUROC                        & AUROC                        & AUROC                        & Acc                          & BL4                          & CDr                          & Acc                          & Acc                          & Acc                          \\ \midrule
1                    & \cmark                & \multicolumn{1}{l}{}  & \multicolumn{1}{l}{} & 66.1                         & 79.1                         & 81.1                         & 77.2                         & 6.9                          & 11.1                         & 57.5                         & 79.5                         & 70.8                         \\
2                    & \multicolumn{1}{l}{}  & \cmark                & \multicolumn{1}{l}{} & 56.9                         & 83.0                         & 85.8                         & 77.5                         & 10.0                         & 18.2                         & 23.5                         & 82.8                         & 59.3                         \\
3                    & \cmark                & \cmark                & \multicolumn{1}{l}{} & 74.5                         & 87.2                         & 88.4                         & \cellcolor[HTML]{EFEFEF}78.3 & 9.9                          & 17.1                         & \cellcolor[HTML]{EFEFEF}67.0 & \cellcolor[HTML]{EFEFEF}84.6 & \cellcolor[HTML]{EFEFEF}77.7 \\ \midrule
4                    & \multicolumn{1}{l}{}  & \multicolumn{1}{l}{}  & \cmark               & \cellcolor[HTML]{EFEFEF}88.0 & \cellcolor[HTML]{EFEFEF}88.9 & \cellcolor[HTML]{EFEFEF}89.3 & 76.5                         & \cellcolor[HTML]{EFEFEF}10.3 & \cellcolor[HTML]{EFEFEF}19.0 & 64.8                         & 81.0                         & 74.6                         \\ \midrule
5                    & \cmark                & \cmark                & \cmark               & \cellcolor[HTML]{C0C0C0}88.7 & \cellcolor[HTML]{C0C0C0}89.0 & \cellcolor[HTML]{C0C0C0}90.1 & \cellcolor[HTML]{C0C0C0}80.0 & \cellcolor[HTML]{C0C0C0}10.7 & \cellcolor[HTML]{C0C0C0}21.0 & \cellcolor[HTML]{C0C0C0}68.7 & \cellcolor[HTML]{C0C0C0}84.6 & \cellcolor[HTML]{C0C0C0}78.3 \\ \bottomrule
\end{tabular}}
\caption{Ablation studies on the different types of objectives, including the fusion-encoders ones (\textit{i.e.}, masked language modeling (MLM) and image-text matching (ITM)) and the dual-encoders one (\textit{i.e.}, image-text contrast (ITC)). 1\%, 10\%, and 100\% represent the different portion of training data.}
\label{table:ablation_study}
\end{table*}

\section{Conclusion}
\label{sec:conclusion}
In this paper, we proposed a simple yet effective scheme to take advantage of both fusion encoders, and dual encoders, where visual and textual prompt pools are used to make our model compatible with different kinds of inputs (\textit{i.e.}, image-only, text-only, and image-text-pair), and thus different types of objectives (\textit{e.g.}, MLM and ITM for fusion-encoders and ITC for dual-encoders) can be adopted for pre-training.
It is worth noting that our proposed approach is complementary to most of the existing Med-VLP models.
Experimental results confirm the validity of our approach, where it achieves state-of-the-art performance on the downstream tasks.
The further analyses investigate the effects of different types of objectives.
Such empirical studies might provide a valuable reference for future research in this area.
{\small
\bibliographystyle{plain}
\bibliography{main}
}
\clearpage
\appendix
\section{More Details of Downstream Evaluation}
\label{sec:downstream-evaluation}
In this section, we detail the descriptions for each downstream evaluation dataset.

\paragraph{CheXpert}
This dataset contains 224,316 chest radiographs labeled for 14 medical observations.
Following the previous studies (GLoRIA \cite{huang2021gloria}), we only keep those front-view radiographs, hold out the expert-labeled validation set as the test set, and randomly sample 5,000 images from the training data for validation.

\paragraph{RNAS Pneumonia}
This dataset consists of 30,000 front-view chest radiographs labeled by ``pneumothorax negative'' or ``pneumothorax positive''.
Following the previous studies (GLoRIA \cite{huang2021gloria}), the train/validation/test split constitutes 70\%/15\%/15\% of the dataset, respectively.

\paragraph{RadNLI}
This dataset contains 19k sentence pairs labeled by ``Entailment'', ``Neutral'', or ``Contradiction''.
We follow IFCC \cite{miura2021ifcc} to produce and pre-process the dataset, which contains the training data from an extra NLI dataset (\textit{i.e.}, MedNLI).

\paragraph{ROCO}
This dataset contains 81k image-text pairs.
For the training and validation set, we adopt the official ones.
For the test procedure, we sample 2,000 pairs from the test set and evaluate the models on the 2,000 pairs to obtain the Recall@K scores.

\paragraph{MIMIC-CXR}
This dataset contains 377,110 chest X-rays.
Different from the pre-training, for downstream evaluations (\textit{i.e.}, text summarization and image-to-text generation), we only keep those front-view x-rays with both the findings and impression section.

\paragraph{VQA-RAD}
This dataset consists of 315 images and 3,515 questions.
We adopt the commonly used version pre-processed by MEVF \cite{nguyen2019mevf}.

\paragraph{SLAKE}
This dataset contains 642 images and 14,028 questions.
We follow the original SLAKE paper \cite{liu2021slake} to prepare and pre-process the dataset and adopt the official dataset split.

\paragraph{MedVQA-2019}
This dataset contains 4,200 images and 15,292 questions.
We follow previous studies \cite{sharma2021medfusenet} to prepare and pre-process the dataset by keeping the main three categories of questions: Modality, Plane, and Organ system.

\end{document}